\pdfoutput=1

\documentclass[11pt]{article}

\usepackage[final]{coling}

\usepackage{times}
\usepackage{latexsym}

\usepackage[T1]{fontenc}

\usepackage[utf8]{inputenc}

\usepackage{microtype}

\usepackage{inconsolata}

\usepackage{graphicx}

\usepackage{amssymb}
\usepackage{array}
 \usepackage{booktabs}
 \usepackage{multirow}
\usepackage{float}
\usepackage{CJKutf8}
\usepackage{pifont}
\usepackage{hyperref}

\title{SCCD: A Session-based Dataset for Chinese Cyberbullying Detection}

\author{
 \textbf{Qingpo Yang},
 \textbf{Yakai Chen},
 \textbf{Zihui Xu},
 \textbf{Yu-ming Shang},
 \textbf{Sanchuan Guo},
 \textbf{Xi Zhang}\thanks{Corresponding author}
\\
 Beijing University of Posts and Telecommunications
\\
\texttt{zhangx@bupt.edu.cn}}

\begin{document}
\maketitle

\begin{abstract}
The rampant spread of cyberbullying content poses a growing threat to societal well-being. However, research on cyberbullying detection in Chinese remains underdeveloped, primarily due to the lack of comprehensive and reliable datasets. Notably, no existing Chinese dataset is specifically tailored for cyberbullying detection. Moreover, while comments play a crucial role within sessions, current session-based datasets often lack detailed, fine-grained annotations at the comment level. To address these limitations, we present a novel Chinese cyberbullying dataset, termed \textbf{SCCD}, which consists of 677 session-level samples sourced from a major social media platform Weibo. Moreover, each comment within the sessions is annotated with fine-grained labels rather than conventional binary class labels. Empirically, we evaluate the performance of various baseline methods on \textbf{SCCD}, highlighting the challenges for effective Chinese cyberbullying detection.\footnote{The proposed SCCD is released in \url{https://github.com/STAIR-BUPT/SCCD}.}

\end{abstract}

\section{Introduction}

The rapid proliferation of social media platforms has exacerbated the severity of cyberbullying. Cyberbullying encompasses various forms of bullying or harassment conducted via digital devices and the Internet, where individuals can view, comment, and share content \cite{CyberbullyDefi}. In response to the rapid growth of harmful content on social media, increasing research efforts have been devoted for automatic cyberbullying detection across various languages \cite{Cheng2019XBullyCD,TGBully,Murshed2022DEARNNAH,Maity2022AMF}, aiming to curb abusive behaviors and prevent further harm \cite{9,10}.

Existing work on cyberbullying detection predominantly concentrates on analyzing isolated social media posts or comments. These sentence-level detection methods commonly focus on text analysis to identify aggressive and harassing contents. For example, \citet{TF-IDF} augmented a unigram model with sentiment coherence attributes, integrating TF-IDF values as content characteristics to enrich the feature set. However, relying solely on sentence-level content features would constrain the ability to capture intricate contextualization and diversity \cite{77}, which is of prominent importance for real-world cyberbully detection. However, a broader session-level analysis is rarely touched in previous studies~\cite{Yi2023LearningLH}.

\begin{figure}[t]
  \includegraphics[width=0.95\columnwidth]{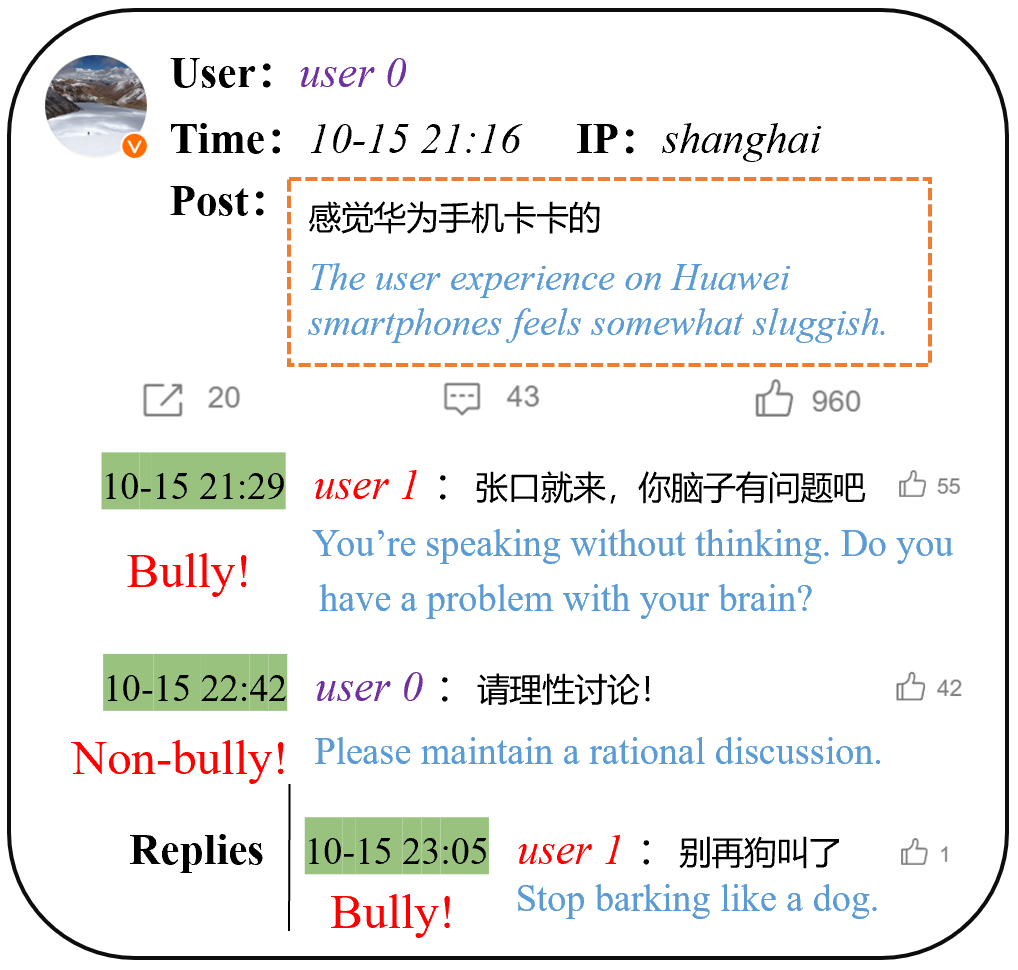}
  \caption{An illustration of a cyberbullying session from Weibo with repetitive offensive behaviors.}
  \label{fig1:session}
\end{figure}


Social media sessions are ubiquitous ecosystems of cyberbullying, which comprise a source post, the subsequent series of comments and associated attributes (e.g., post time, location and the number of likes) \cite{Yi2022SessionbasedCD}. A key characteristic of cyberbullying is the repeated aggression nature \cite{CB-repeat}. Figure \ref{fig1:session} shows an example of a cyberbullying session. The repetitiveness cannot be captured by previous sentence-level methods, which motivates researchers to move to session-level cyberbullying detection. Despite recent advancements on session-level studies, they focus on English~\cite{CBbias, Yi2023LearningLH}, while the research on other languages is insufficient.


\begin{table}[t]
\footnotesize
\centering
\begin{tabular}{m{1.5cm} m{1.4cm} m{1.3cm} m{1.4cm}}
    \toprule
    \textbf{Dataset} & \textbf{Session level} & \textbf{Size} & \textbf{Balance index}  \\

    \midrule
    
    \citet{2015Instagram} & Yes & 2218 & 29\%  \\
    \hline
    \citet{2015Vine} & Yes & 970 & 31\%   \\
    \hline
    \citet{32} & Yes & 115864 & 11\%   \\
    \hline
    \citet{34} & No & 47000 & 16\%   \\
    \bottomrule
\end{tabular}
\caption{The imbalance of available cyberbullying datasets. If the dataset is session-based, size denotes the number of sessions. Otherwise, it indicates the number of texts. Balance index denotes the proportion of cyberbullying instances within the dataset.}
\label{table1}
\end{table}

Cyberbullying datasets are fundamental to the development of effective detection models. However, there only exists a handful of English session-level datesets, lacking Chinese datasets. Moreover, existing datasets generally suffer from several limitations. (1) The class imbalance existing at both session and sentence levels \cite{Yi2022SessionbasedCD} would degrade the performance of machine learning classifiers \cite{imblance1,imblance2}. Table \ref{table1} highlights the imbalance factor in existing datasets; (2) Only the overall session-level label is provided, and the lack of fine-grained labels of comments cannot support reliable and trustworthy prediction.

To address these gaps, we introduce the first publicly available Chinese dataset for cyberbullying detection: SCCD (\textbf{S}ession-based \textbf{C}hinese \textbf{C}yberbullying \textbf{D}ataset). SCCD is balanced and contains 677 sessions, with 52.3\% classified as instances of cyberbullying. Each cyberbullying session is annotated with an overall severity level categorized as low, medium, or high. All comments are carefully annotated, providing detailed labels that capture multiple aspects of the text. Examples of comments with fine-grained labels are shown in Table \ref{table_example}. In particular, the dataset offers the source post, the comments, user details and other relevant attributes. For further details, please refer to Appendix \ref{sec:A}.

\begin{table*}
  \small
  \centering
  \renewcommand\arraystretch{1.2}
  \begin{tabular}{  m{0.32\textwidth}  m{0.1\textwidth}  m{0.08\textwidth}  m{0.08\textwidth} m{0.08\textwidth} m{0.08\textwidth}}
    \toprule

     \textbf{Comment} & \textbf{Cyberbully} & \textbf{Expression} & \textbf{Sarcastic} & \textbf{Target}  & \textbf{Group Category} \\
    \midrule
          \begin{CJK*}{UTF8}{gbsn} 河南人看到井盖就走不动了。  \end{CJK*}
          & \multirow{2}{*}{\shortstack{\\ \\ \\CB}} & \multirow{2}{*}{\shortstack{\\ \\ \\Implicit}} & \multirow{2}{*}{\shortstack{\\ \\ \\Yes}} & \multirow{2}{*}{\shortstack{\\ \\ \\Group}} & \multirow{2}{*}{\shortstack{\\ \\ \\Region}} \\

          When Henan people see a manhole cover, they just can't resist stopping. & & & & \\
          
    \hline
         \begin{CJK*}{UTF8}{gbsn}  你个废物闭嘴吧！ \end{CJK*}
         & \multirow{2}{*}{CB} & \multirow{2}{*}{Explicit} & \multirow{2}{*}{No} & \multirow{2}{*}{Individual} & \multirow{2}{*}{-} \\

        Shut up, you loser! & & & & \\
         
    \hline

        \begin{CJK*}{UTF8}{gbsn} 妈的，我们输了！  \end{CJK*}
        & \multirow{2}{*}{Non-CB} & \multirow{2}{*}{-} & \multirow{2}{*}{-} & \multirow{2}{*}{-} & \multirow{2}{*}{-}  \\

        Damn, we lost. & & & & \\
        
    \bottomrule
  \end{tabular}
  \caption{\label{table_example}
    Three types of comments in SCCD: group-targeted cyberbullying comments, individual-targeted cyberbullying comments and normal comments. CB refers to cyberbullying.
  }
\end{table*}

 The key contributions of our work are summarized as follows:
\begin{itemize}
    \item To the best of our knowledge, SCCD is the first open-source Chinese cyberbullying detection dataset, which systematically gathers and formalizes sessions from diverse topics.
    \item SCCD also presents fine-grained annotations of session comments to enable more detailed analysis and more explainable detection. 
   
    \item Experimental validation with several established baselines on SCCD identifies the challenges for future research on Chinese cyberbullying detection.
\end{itemize}

\begin{table*}
  \small
  \centering
  \renewcommand\arraystretch{1.2}
  \begin{tabular}{m{0.2\textwidth} m{0.08\textwidth}<{\centering}  m{0.18\textwidth}<{\centering} m{0.1\textwidth}<{\centering} m{0.7cm}<{\centering}
  m{0.16\textwidth}<{\centering}}
    \toprule
    \textbf{Dataset} & \textbf{Size} & \textbf{Expression Category} & \textbf{Sarcasm} & \textbf{Target}  & \textbf{Group Category}\\
    \midrule
        COLD \cite{COLD} & 37480  & \ding{56} & \ding{56} & \ding{52} & \ding{56} \\
    \hline
        SWSR \cite{SwSR} & 8969  & \ding{52} & \ding{56} & \ding{52} & \ding{56} \\
    \hline
        TOXICN \cite{TOXIGEN} & 12011  & \ding{52} & \ding{56} & \ding{56} & \ding{52}\\
    \hline
        SCCD (ours) & 38999 & \ding{52} & \ding{52} & \ding{52} & \ding{52} \\
    \bottomrule
  \end{tabular}
  \caption{\label{table2}
    Comparison between proposed dataset and other related Chinese datasets. The expression category includes explicit expression and implicit expression.
  }
\end{table*}

\section{Related Work}
\subsection{Cyberbullying Detection}
Recently, most researchers have utilized methods of deep learning to tackle the problem of cyberbullying detection. \citet{HANCD} and \citet{HENIN} used a hierarchical network to model the structure of social media sessions and applied an attention mechanism to capture multi-grained embeddings. More recent research turned into investigating temporal information of cyberbullying \cite{Time1,Time2,InstagramTemporal}. For example, \citet{TGBully} attempted to utilize the interactions of users themselves within a session by modelling topic coherence and temporal user interactions.

In addition to text, numerous methods incorporated multimodal information into cyberbullying detection. XBully \cite{Cheng2019XBullyCD} reorganized multimodal social media data into a heterogeneous network. \citet{Maity2022AMF} introduced the task of sentiment-emotion-sarcasm aware multimodal cyberbully detection and proposed an attention based multi-task multimodal framework.

In conclusion, it is evident that research on cyberbullying detection in foreign countries has reached a relatively mature stage. In contrast, this field remains largely unexplored within China, partly due to the lack of available Chinese datasets.

\subsection{Cyberbullying Dataset}
\subsubsection{Non-Chinese Dataset}

Efforts to build non-Chinese cyberbullying datasets at the sentence level are comprehensive \cite{28,34,Maity2022AMF}, but our focus is on session-based datasets. Two session-based datasets \cite{2015Instagram,2015Vine} labeled each session as either cyberbully or non-cyberbully. \citet{InstagramTemporal} extracted 100 sessions and manually labeled each comment to study its temporal properties. Later, \citet{MulitiLabelsIns} expanded the labels to capture diverse granularities, such as purpose, to explore content patterns. These studies demonstrate that researchers have increasingly focused on the analysis of comments within sessions.

Nevertheless, current session-based datasets often lack detailed comment labels or are too small for extensive research. Hence, our dataset provides fine-grained annotations for all comments within each session.

\subsubsection{Chinese Dataset}
There remains a dire scarcity of relevant dataset in Chinese. To the best of our knowledge, there is no available session-based Chinese dataset for cyberbullying detection. In Table \ref{table2}, we list all relevant sentence-level datasets in Chinese to compare with ours. \citet{SwSR} presented the first Chinese sexism dataset as well as a large Chinese lexicon and \citet{COLD} proposed the first benchmark– COLD for Chinese offensive language analysis. Previous work failed to separate hate speech from general offensive language, so \citet{TOXICN} proposed a fine-grained dataset of Chinese toxic language with an insult lexicon.

However, they are not specifically designed for Chinese cyberbullying. In addition, the available datasets are not session-based, lacking the necessary contextual information for analysis. Therefore, our work presents the first Chinese cyberbullying dataset with fine-grained analysis to fill these gaps.

\section{Data Construction}

\subsection{Overview}
\begin{figure*}
  \centering
  \includegraphics[width=2\columnwidth]{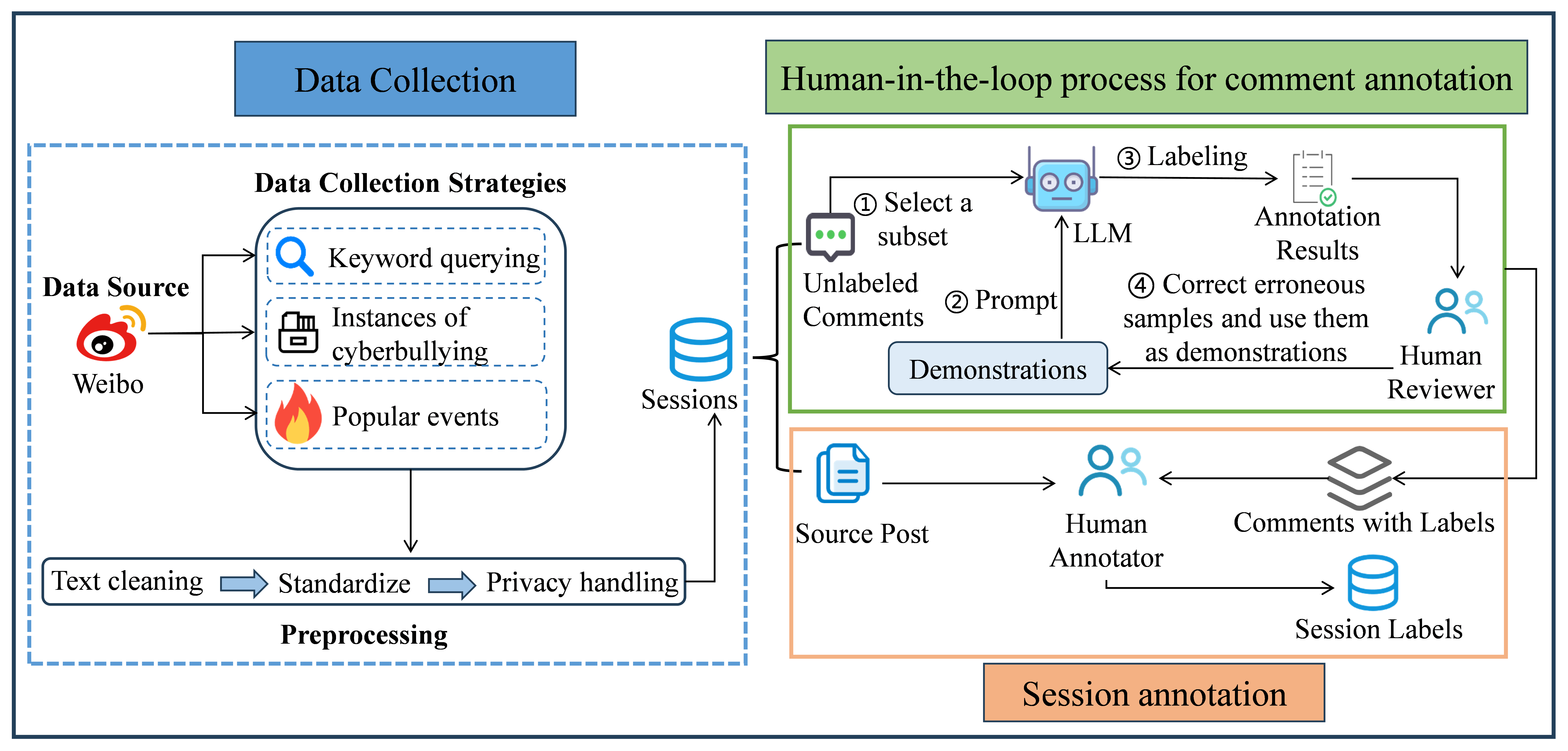}
  \caption{Overview of data construction. The data annotation process involves two steps: first, using a large language model (LLM) to annotate the comments, and then manually labeling the conversations.}
  \label{fig:overview}
\end{figure*}

In this section, we describe the annotation strategies employed and the construction of SCCD, which is divided into four stages: data collection, data preprocessing, data annotation and data validation. An overview of data construction is shown in Figure \ref{fig:overview}. Finally, a snapshot of basic statistics of the final dataset is shown.

\subsection{Data Collection and Preprocessing}

In order to gain insights into the current status of cyberbullying in China, we crawl the published sessions from \textit{Weibo}, a public online social platform that serves as a prototypical representative of Chinese social media due to its vast user base composed predominantly of local individuals. Our data collection employs two strategies: keyword querying and crawling from typical instances of cyberbullying. This approach ensures that our dataset captures both prevalent topics related to cyberbullying as well as specific, high-profile cases that offer valuable insights into the phenomenon.

We find that cyberbully tends to occur in discussions of several sensitive topics, including \textit{gender}, \textit{region}, \textit{race}, and \textit{LGBTQ}. Therefore, we compile a list of relevant keywords for each topic and use them to obtain related samples. The collected keywords are shown in Appendix \ref{sec:keywords}. Subsequently, we manually compile a list of prominent cyberbullying incidents in China over the past three years and crawl data related to these events. In addition, to ensure the representativeness and universality of our dataset, we also crawl data from daily popular events across various themes, including entertainment, society, and politics. The overview of the data distribution associated with the collection strategy is presented in Appendix \ref{sec:source}.

User-generated content naturally contains a high level of noise. To minimize the noise, we apply various preprocessing steps to normalize the noisy posts and comments. We clean the noisy information in the original text, including URLs, emojis, white space and some irrelevant contents, such as "retweeted Weibo posts." Meanwhile, we standardize the text by converting all letters to lowercase and transforming traditional Chinese characters into simplified Chinese characters. To protect user privacy, we anonymize the data, by removing all \textit{@USERs} from the text and encrypting the \textit{IDs}.

\subsection{Data Annotation}

\begin{figure}
  \includegraphics[width=\columnwidth]{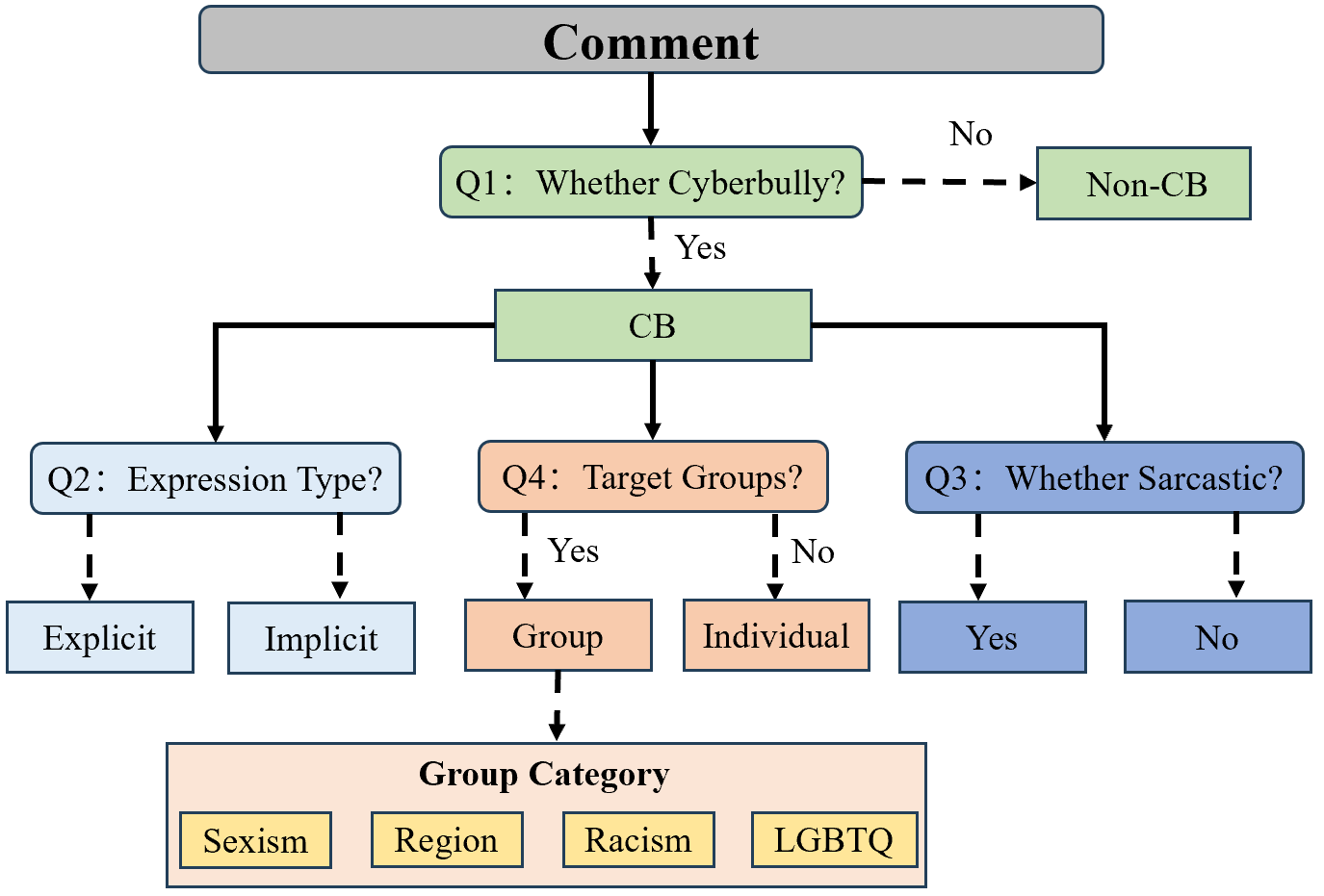}
  \caption{The annotation guideline of comments. It outlines four key questions that determine the five labels.}
  \label{fig:guide}
\end{figure}

We have established a standardized annotation guide to assist annotators in the fine-grained labeling of comments, which is shown in Figure \ref{fig:guide}.

\subsubsection{Annotation Procedure}

\citet{GPTAnno} demonstrated the great potential of ChatGPT as a data annotation tool, due to its better performance than human in detecting implicit hateful speech and providing natural language explanations. Cyberbullying and hate speech, both categorized under toxic language, share numerous similarities, particularly in their implicit expressions, which pose significant challenges for detection. Therefore, to improve the annotation efficiency, we introduce the large language model (LLM) to facilitate the labeling process. Considering the annotation requirements in the Chinese context and after conducting evaluations, we utilize Doubao-pro-128k as the annotation tool. The annotation procedure consists of two distinct stages: comment annotation and session annotation.

\textbf{Comment Annotation}. We propose a human-in-the-loop approach to enhance the collaboration between human annotators and the LLM. Initially, the session’s post is provided to the LLM to establish context and background understanding. We then employ demonstration-based prompting to enable few-shot learning in the LLM. Subsequently, we engage a human-in-the-loop process: select a subset of comments, label them with the LLM, verify these labels by human annotators, manually rectify any erroneous samples and utilize them as demonstrations, pass demonstrations to the LLM. In Figure \ref{fig:overview} we present the pipeline of our methodology.

Specifically, within the loop, we systematically select comments in descending order of likes, prioritizing those with a higher degree of popularity. Human annotators are tasked with verifying the labels generated by the LLM. If the accuracy of the labels reaches 90\%, the LLM is deemed a competent annotator, and the remaining comments in this session are subsequently labeled directly by the LLM. Otherwise, the process will continue in a loop until the threshold is met.

\textbf{Session Annotation}. Once all comments in a session are annotated, human annotators review the initial post along with the annotated comments to determine whether the session involves cyberbullying or remains normal. If a session is identified as cyberbullying, human annotators will assess the cyberbullying severity of the session.

\subsubsection{LLM Annotation}
To optimize the LLM’s performance as an annotator, we employ prompt engineering to equip it with the necessary knowledge and enhance its ability to understand conversational contexts.

\textbf{Role Definition}. Defining explicit roles for large language models (LLMs) is one of the most prevalent approaches in prompt engineering, significantly improving the quality and efficiency of their responses. To leverage the extensive knowledge encoded within large language models, we define the LLM as a Chinese cyberbully specialist and a social media veteran. The specific prompt is presented in Appendix \ref{sec:role}.


\textbf{Demonstration-based prompting}. As a method of few-shot learning, demonstration-based prompting, has been shown to activate the in-context learning capabilities of LLMs and guide the models to superior performance \cite{Gao2021,TOXIGEN}. Here, owing to the strong correlation between comments and conversational contexts, we manually select and annotate a subset of comments for each session, providing detailed explanations to clarify the rationale behind each annotation. Specifically, human annotators are instructed to select 10\% of the comments they deemed both challenging and representative for LLMs. Then, the carefully selected comments, along with annotations and explanations, are passed to the LLM, enhancing its annotation capabilities. 

\subsubsection{Human Annotation}
We employed five native Chinese speakers from our team for the labeling tasks, ensuring a gender balance with three male and two female annotators, all of whom possess expertise in cyberbullying research. Each annotator underwent rigorous training and passed an annotation test successfully.

\subsection{Annotation Validation}

As a key quality assurance measure, we sample 10\% of the comments from each session and review them for labeling accuracy. If the accuracy falls below 90\%, we manually reannotate all comments within that session. We find that 9\% of the sessions failed to meet the required accuracy, with the lowest accuracy being 81\%. Among the errors, expression-related issues are the most prevalent.


    

\begin{table}
  \footnotesize
  \centering
  \renewcommand\arraystretch{1.1}
  \begin{tabular}{ m{0.09\textwidth}  m{0.05\textwidth}<{\centering}  m{0.03\textwidth}<{\centering}  m{0.06\textwidth}<{\centering}  m{0.03\textwidth}<{\centering} m{0.06\textwidth}<{\centering}}
    \toprule

    & \textbf{Total} & \textbf{Low} & \textbf{Medium} & \textbf{High} & \textbf{Avg. L}  \\
    \midrule
        CB Session & 354 (52.3\%) & 106 & 105 & 143 & 67.9\\
    \midrule
        Avg. CB & \textit{/} & 20.6 & 20.9 & 34.9  & \textit{/}\\
    \bottomrule
  \end{tabular}
  \caption{\label{table3}
    The label distribution of the sessions in SCCD. Cyberbullying sessions constitute 52.3\% of the dataset. CB refers to cyberbullying and Avg. L is the average number of comments per session. Avg. CB denotes the average number of cybeybullying comments per session.
  }
\end{table}

\begin{table*}
  \small
  \centering
  \renewcommand\arraystretch{1.1}
  \begin{tabular}{ m{0.07\textwidth} 
     m{0.12\textwidth}<{\centering} 
   m{0.03\textwidth}<{\centering}   m{0.03\textwidth}<{\centering}  
   m{0.03\textwidth}<{\centering}   m{0.03\textwidth}<{\centering}  
   m{0.03\textwidth}<{\centering}   m{0.03\textwidth}<{\centering}  
   m{0.03\textwidth}<{\centering}   m{0.03\textwidth}<{\centering}  
   m{0.03\textwidth}<{\centering}   m{0.07\textwidth}<{\centering} }
    \toprule
     & \multirow{2}{*}{\shortstack{ \\ \\  \textbf{CB} \\ \textbf{Comments}}} 
     & \multicolumn{2}{c}{\textbf{Expression}} 
     & \multicolumn{2}{c}{\textbf{Sarcasm}} 
     & \multicolumn{2}{c}{\textbf{Target}} 
     & \multicolumn{4}{c}{\textbf{Group Category}}\\
    \cmidrule(lr){3-12}
    & & Exp. & Imp. & Yes & No & Ind. & Grp. & Sex. & Reg. & Rac. & LGBTQ\\
    \midrule
        \shortstack{CB \\ Session} & 9380 (39\%) & 8262 & 1118 & 1270 & 8110 & 5593 & 3787 & 578 & 1390 & 1349 & 470 \\
    
        Non-CB \quad Session & 425 (2.8\%) & 371 & 54 & 56 & 369 & 269 & 156 & 30 & 90 & 29 & 7 \\

        Total & 9805 (25.1\%) & 8633 & 1172 & 1326 & 8479 & 5862 & 3943 & 608 & 1480 & 1378 & 477 \\
        
    \bottomrule
  \end{tabular}
  \caption{\label{table4}
    The basic statistics of SCCD (Exp.: Explicit, Imp.: Implicit, Ind.: Individual, Grp.: Group, Sex.: Sexism, Reg.: Region, Rac.: Racism). In the column "CB Comments," the values in parentheses indicate the proportion of cyberbullying comments to the total number of comments.
  }
\end{table*}

\subsection{Data Statistics}
Table \ref{table3} shows the label distribution of the sessions. The dataset consists of a total of 677 sessions, where 354 are tagged as cyberbullying (further labeled with cyberbullying severity). As it can be seen, our dataset is balanced. Table \ref{table4} presents basic statistics of the final comments. Across all 677 sessions, 9,805 comments are labeled as cyberbullying and 29,194 as non-cyberbullying.

We observe an imbalance in sample distribution across different categories of comments, with notably fewer cyberbullying comments compared to normal ones. This distribution mirrors the real-world conditions of social media platforms \cite{DataImbalance}. Therefore, we opt not to implement additional interventions to address the imbalance.

Next, we seek to elucidate the relationship between expression categories and sarcasm. Our corpus contains 1,172 cyberbullying comments with implicit expressions, of which 318 are marked as sarcastic, while 854 are non-sarcastic. In contrast, from a total of 8,633 comments with explicit expressions, only 1,008 are designated as sarcastic, whereas the overwhelming majority, 7,625, are labeled as non-sarcastic. It indicates that implicit cyberbullying comments are more likely to exhibit sarcasm than explicit comments.

\section{Experiments}
To illustrate the complexity of cyberbullying detection at hand, we present initial experimental results on the novel dataset, which are intended to serve as benchmarks for further experiments. Since our dataset contains labels for both sessions and individual comments, the experiments are conducted in two parts:

\begin{itemize}
    \item \textbf{CL-CD (Comment-Level Cyberbullying Detection):} We evaluate the performance of several models in recognizing cyberbullying instances at the sentence-level, which is to assign the label y (CB or Non-CB) to the given comment c.
    \item \textbf{SL-CD (Session-Level Cyberbullying Detection):} We present the results of session-based cyberbullying detection methods on our dataset.
\end{itemize}

\subsection{Baselines}

Here we introduce all baseline models of our experiments.

\textbf{Baidu Text Censor (Baidu TC)}\footnote{\url{https://ai.baidu.com/tech/textcensoring}}. As a widely used online API, it is designed to detect and filter harmful and inappropriate content across various online platforms.

\textbf{COLDETECTOR} \cite{COLD}. As an offensive language detecting model based on bert-base-chinese, it is fine-tuned on the proposed COLDataset.

\textbf{CNN} \cite{CNN}. CNN is a type of feedforward neural network that incorporates convolutional computations and possesses a deep structure, widely utilized in single sentence classification.

\textbf{LSTM} \cite{LSTM}. LSTM is a special type of Recurrent Neural Network (RNN) that effectively resolves long-sequence dependency via memory cells and gates.

\textbf{BERT} \cite{BERT}. Bert is a language representation model designed to pre-train deep bi-directional representations from unlabeled text. The version of bert-base-chinese,\footnote{\url{https://huggingface.co/bert-base-chinese}} which has 12 layers and 12 attention heads, is used as the baseline.

\textbf{RoBERTa} \cite{RoBERTa}. RoBERTa is an optimised BERT-based model, which removes Next Sentence Prediction (NSP), employs larger datasets, longer training, and dynamic masking. Similarly, we utilize the most commonly used chinese version of Roberta, roberta-base-chinese.\footnote{\url{https://huggingface.co/hfl/chinese-roberta-wwm-ext}}

\textbf{GPT-4} \cite{GPT-4}. We use the version of GPT-4O. Due to the extensive length of sessions, conveying all comments and supplementary information to GPT is both economically burdensome and inefficient. Hence, we provide only the post content and the five most-liked comments for analysis by GPT.

\begin{table}
  \small
  \centering
  \renewcommand\arraystretch{1.2}
  \begin{tabular}{  m{0.12\textwidth}<{\centering}  m{0.08\textwidth}<{\centering}  m{0.08\textwidth}<{\centering}  m{0.08\textwidth}<{\centering}}
    \toprule

    \rule{0pt}{10pt} 
     \textbf{Model} & \textbf{Precision} & \textbf{Recall} & \textbf{Micro F1}  \\
    \midrule
         Baidu TC & 62.3 & 21.9 & 76.7 \\

         COLDETECTOR & 64.0 & 39.0 & 79.2 \\

        Bert & $ 70.2\pm4.2 $  & $62.5 \pm 5.7$ & $84.2 \pm 0.8$ \\

        Roberta & $\textbf{73.5} \pm \textbf{2.7}$ & $\textbf{65.9} \pm \textbf{5.7}$  & $\textbf{85.8} \pm \textbf{0.2}$ \\

    \bottomrule
  \end{tabular}
  \caption{\label{table5}
    Results of various models on our dataset at the sentence-level. The best results are in \textbf{bold}.
  }
\end{table}

\begin{table*}
  \small
  \centering
  \renewcommand\arraystretch{1.4}
  \begin{tabular}{ m{0.2\textwidth}<{\centering}  m{0.08\textwidth}<{\centering}  m{0.12\textwidth}<{\centering}  m{0.12\textwidth}<{\centering}  m{0.12\textwidth}<{\centering}}
    \toprule

    \rule{0pt}{10pt} 
     \textbf{Approach} & \textbf{Model} & \textbf{Precision} & \textbf{Recall} & \textbf{Micro F1}  \\
    \midrule
        \multirow{2}{*}{\shortstack{Neural text \\ classification models}} & CNN & $86.4 \pm 9.9$ & $63.8 \pm 12.4$ & $72.0 \pm 2.8$ \\

         & LSTM & $82.9 \pm 1.7$ & $63.3 \pm 10.6$  & $71.7 \pm 4.3$  \\

        \multirow{2}{*}{\shortstack{Pre-trained \\ language models}} & Bert & $89.5 \pm 7.5$ & $84.7 \pm 8.8$  & $84.9 \pm 2.3$  \\

        & Roberta & $\textbf{91.9} \pm \textbf{3.7}$  & $83.4 \pm 2.1$  & $86.3 \pm 1.8$  \\

        \shortstack{Large \\ Language models} & GPT-4 & 91.4 & \textbf{89.0} & \textbf{90.0} \\
        
    \bottomrule
  \end{tabular}
  \caption{\label{table6}
    Evaluation of three types of session-based cyberbullying detection models. The best results in each group are shown in \textbf{bold}.
  }
\end{table*}

\subsection{Experimental Setup}
For the two experiments, different baseline models are employed.

In \textbf{CL-CD}, four existing methods are evaluated: Baidu TC, COLDETECTOR, BERT and RoBERTa. BERT and RoBERTa are fine-tuned on the training data to optimize model performance.

For \textbf{SL-CD}, we utilize three types of baseline models: neural text classification models (CNN, LSTM), several transformer-based pre-trained language models (BERT, RoBerta) and a large language model (GPT-4).

CNN is often used for single sentence classification. In our experiment, we implement it at the comment level, averaging the resulting comment representations to derive a session-level representation. Likewise, we employ LSTM to model the comments and classify sessions by averaging their comment-level representations. In addition, PLMs are limited in the length of the text inputs they can handle (usually 512 tokens), while the limited length is not enough for social media sessions, which poses a challenge for modelling lengthy social media sessions for cyberbullying detection. Therefore, we utilize the truncation strategy defined by \citet{transaction}. 

\textbf{Implementation Details}. We employ three widely recognized evaluation metrics in cyberbullying detection tasks: recall (\textit{R}), precision (\textit{P}) and micro-F1 (\textit{Mic F1}), which are also typically used in imbalanced classification tasks. All the samples in SCCD are split into a training set and a test set with a ratio of 7:3. All baselines, except for Baidu Text Censor and COLDETECTOR, are repeated five times, with average performance and standard deviation reported.

\subsection{Experimental Results}

\subsubsection{CL-CD}

The experimental results are shown in Table \ref{table5}. Analysis of the experimental results leads to the following conclusions:

(1) Achieving satisfactory performance on this task solely with existing resources is difficult. To investigate whether Chinese cyberbullying texts can be effectively detected by current resources alone, we evaluate Baidu TC and COLDETECTOR. However, they perform poorly on our dataset, with recall scores of only 0.219 and 0.39. The low recall indicates that the models frequently fail to detect actual instances of cyberbullying, resulting in a high rate of missed identifications. This suggests significant limitations in handling cyberbullying texts that are subtle, ambiguous, or implicitly expressed.

(2) Our dataset facilitates the advancement of Chinese cyberbullying detection in online communities. The fine-tuned BERT and RoBERTa models demonstrate exceptional performance, significantly outperforming other models with an average recall score improvement of 33.75\%.

(3) Despite fine-tuning, the models still show limited effectiveness in discovering the cyberbullying contents, often recognizing cyberbullying texts as \textit{Non-CB}. We hypothesize that the low recall may be attributed to a lack of contextual information.

\begin{table}
  \small
  \centering
  \renewcommand\arraystretch{1.2}
  \begin{tabular}{  m{0.17\textwidth}<{\centering}  m{0.07\textwidth}<{\centering}  m{0.05\textwidth}<{\centering}  m{0.08\textwidth}<{\centering}}
    \toprule

    \rule{0pt}{6pt} 
     \textbf{Approach} & \textbf{Precision} & \textbf{Recall} & \textbf{Micro F1}  \\
    \midrule
         Most liked comments & 91.4 & \textbf{89.0} & \textbf{90.0} \\

         Without comments & 90.4 & 64.5 & 77.9 \\

         Random comments & \textbf{92.0}  & 78.6 & 85.3\\

    \bottomrule
  \end{tabular}
  \caption{\label{table7}
    Performance of GPT with different comment selection strategies. The best results are in \textbf{bold}.
  }
\end{table}

\subsubsection{SL-CD}

This set of experiments seeks to show the performance of existing session-based models when utilized within Chinese linguistic contexts. According to \citet{TGBully} and \citet{Yi2023LearningLH}, three categories of baselines are used to be evaluated on the dataset. The results are reported in Table \ref{table6}. From the table, we can find that:

(1) Compared with traditional neural text classification models,  the pre-trained language models achieve better performance, even when provided with a limited subset of comments. The superior performance of PLMs can be attributed to their advanced feature extraction capabilities and the comprehensive language understanding gained through pre-training.

(2) As expected, GPT achieves the best performance across all evaluation metrics, particularly excelling with an exceptionally high recall (89\%), which indicates that large language models hold great potential for cyberbullying detection at the session level.

(3) While all models achieve high precision, their recall rates are relatively lower in comparison. For example, CNN achieves notable performance in precision (86.4\%), but its recall rate is only 63.8\%.

Finally, we conduct additional experiments to explore the significance of comments in session-based cyberbullying detection. We use GPT as the baseline model, maintaining the experimental setup used in \textbf{SL-CD}. Additionally, we design two control experiments: one without comments and another with randomly selected comments. The results are presented in Table \ref{table7}. Without comments, GPT demonstrates a low recall (64.5\%), significantly underperforming compared to when comments are included. This highlights the importance of comments in session contexts. In addition, compared to randomly selected comments, using the strategy of selecting comments with the most likes improves the recall score by 10.4\%. The improvement means that highly liked comments tend to be more representative and contain richer information, enhancing the model’s ability to comprehend context and detect cyberbullying more effectively.

\section{Conclusion}

In this paper, we propose SCCD, a Chinese session-based dataset, which represents a pioneering effort for Chinese cyberbullying detection. We annotate the entire corpus of session comments with a fine-grained labeling scheme, which is overlooked by existing session-level datasets. Comments with detailed labels enable diverse research directions in the field of cyberbullying, like studying temporal properties of cyberbullying and mitigating bias. It can also be used as a benchmark for the evaluation of cyberbullying detection models. We evaluate various types of widely used models and reveal that detecting cyberbullying in Chinese contexts is challenging. Additionally, through comparative experiments, we highlight the critical role of comments in enhancing the effectiveness of session-based cyberbullying detection. We expect that our resources, benchmarks, and analyses will assist relevant professionals in detecting cyberbullying.

\section*{Limitation}

The annotations for our comments are primarily generated by a large language model. Although partial manual verification and random sampling checks are conducted to ensure a certain level of accuracy, labeling errors are inevitable in data that have not undergone human review. If all comments were annotated manually, the performance of the existing model would likely improve significantly.

Furthermore, during the data collection, we use a keyword-based query approach focused on specific topics. This method may introduce potential biases into our dataset, such as lexical bias. In future work, we plan to explore ways to mitigate biases in conversation-based cyberbullying datasets.

\section*{Ethical considerations}

During the data collection, we strictly adhere to the terms of service of the relevant platforms. All user-related data undergo rigorous de-identification procedures to ensure that no personally identifiable information is disclosed.

The objective of our research is to detect and safeguard against cyberbullying, rather than to propagate harmful content. The dataset introduced is designated exclusively for academic research and for the development of tools aimed at preventing cyberbullying. Additionally, any aggressive or derogatory content utilized within this paper serves an illustrative function and does not reflect the perspectives of the authors.

\section*{Acknowledgments}

This work is supported by the National Key Research and Development Program (Grant No. 2023YFC3303800).

\bibliography{custom}

\begin{thebibliography}{39}
\providecommand{\natexlab}[1]{#1}

\bibitem[{Achiam et~al.(2023)Achiam, Adler, Agarwal, Ahmad, Akkaya, Aleman, Almeida, Altenschmidt, Altman, Anadkat et~al.}]{GPT-4}
Josh Achiam, Steven Adler, Sandhini Agarwal, Lama Ahmad, Ilge Akkaya, Florencia~Leoni Aleman, Diogo Almeida, Janko Altenschmidt, Sam Altman, Shyamal Anadkat, et~al. 2023.
\newblock Gpt-4 technical report.
\newblock \emph{arXiv preprint arXiv:2303.08774}.

\bibitem[{Alhajji et~al.(2019)Alhajji, Bass, and Dai}]{CyberbullyDefi}
Mohammed Alhajji, Sarah~Bauerle Bass, and Ting Dai. 2019.
\newblock \href {https://api.semanticscholar.org/CorpusID:201098275} {Cyberbullying, mental health, and violence in adolescents and associations with sex and race: Data from the 2015 youth risk behavior survey}.
\newblock \emph{Global Pediatric Health}, 6.

\bibitem[{Chawla(2005)}]{imblance1}
N.~Chawla. 2005.
\newblock \href {https://api.semanticscholar.org/CorpusID:15273230} {Data mining for imbalanced datasets: An overview}.
\newblock In \emph{The Data Mining and Knowledge Discovery Handbook}.

\bibitem[{Chen and te~Li(2020)}]{HENIN}
Hsin-Yu Chen and Cheng te~Li. 2020.
\newblock \href {https://api.semanticscholar.org/CorpusID:222271961} {Henin: Learning heterogeneous neural interaction networks for explainable cyberbullying detection on social media}.
\newblock In \emph{Conference on Empirical Methods in Natural Language Processing}.

\bibitem[{Cheng et~al.(2019{\natexlab{a}})Cheng, Guo, Silva, Hall, and Liu}]{HANCD}
Lu~Cheng, Ruocheng Guo, Yasin~N. Silva, Deborah~L. Hall, and Huan Liu. 2019{\natexlab{a}}.
\newblock \href {https://api.semanticscholar.org/CorpusID:164326774} {Hierarchical attention networks for cyberbullying detection on the instagram social network}.
\newblock In \emph{SDM}.

\bibitem[{Cheng et~al.(2019{\natexlab{b}})Cheng, Li, Silva, Hall, and Liu}]{Cheng2019XBullyCD}
Lu~Cheng, Jundong Li, Yasin~N. Silva, Deborah~L. Hall, and Huan Liu. 2019{\natexlab{b}}.
\newblock \href {https://api.semanticscholar.org/CorpusID:59261205} {Xbully: Cyberbullying detection within a multi-modal context}.
\newblock \emph{Proceedings of the Twelfth ACM International Conference on Web Search and Data Mining}.

\bibitem[{Cheng et~al.(2021)Cheng, Mosallanezhad, Silva, Hall, and Liu}]{CBbias}
Lu~Cheng, Ahmadreza Mosallanezhad, Yasin~N. Silva, Deborah~L. Hall, and Huan Liu. 2021.
\newblock \href {https://api.semanticscholar.org/CorpusID:236460063} {Mitigating bias in session-based cyberbullying detection: A non-compromising approach}.
\newblock In \emph{Annual Meeting of the Association for Computational Linguistics}.

\bibitem[{Cheng et~al.(2020)Cheng, Shu, Wu, Silva, Hall, and Liu}]{Time1}
Lu~Cheng, Kai Shu, Siqi Wu, Yasin~N. Silva, Deborah~L. Hall, and Huan Liu. 2020.
\newblock \href {https://api.semanticscholar.org/CorpusID:221006202} {Unsupervised cyberbullying detection via time-informed gaussian mixture model}.
\newblock \emph{Proceedings of the 29th ACM International Conference on Information \& Knowledge Management}.

\bibitem[{Dadvar et~al.(2014)Dadvar, Trieschnigg, and de~Jong}]{28}
Maral Dadvar, Dolf Trieschnigg, and Franciska de~Jong. 2014.
\newblock Experts and machines against bullies: A hybrid approach to detect cyberbullies.
\newblock In \emph{Advances in Artificial Intelligence}, pages 275--281, Cham. Springer International Publishing.

\bibitem[{Dani et~al.(2017)Dani, Li, and Liu}]{TF-IDF}
Harsh Dani, Jundong Li, and Huan Liu. 2017.
\newblock \href {https://api.semanticscholar.org/CorpusID:40987224} {Sentiment informed cyberbullying detection in social media}.
\newblock In \emph{ECML/PKDD}.

\bibitem[{Devlin et~al.(2019)Devlin, Chang, Lee, and Toutanova}]{BERT}
Jacob Devlin, Ming-Wei Chang, Kenton Lee, and Kristina Toutanova. 2019.
\newblock \href {https://api.semanticscholar.org/CorpusID:52967399} {Bert: Pre-training of deep bidirectional transformers for language understanding}.
\newblock In \emph{North American Chapter of the Association for Computational Linguistics}.

\bibitem[{Gao et~al.(2021)Gao, Fisch, and Chen}]{Gao2021}
Tianyu Gao, Adam Fisch, and Danqi Chen. 2021.
\newblock \href {https://api.semanticscholar.org/CorpusID:229923710} {Making pre-trained language models better few-shot learners}.
\newblock In \emph{Annual Meeting of the Association for Computational Linguistics}.

\bibitem[{Ge et~al.(2020)Ge, Cheng, and Liu}]{TGBully}
Suyu Ge, Lu~Cheng, and Huan Liu. 2020.
\newblock \href {https://api.semanticscholar.org/CorpusID:226227471} {Improving cyberbullying detection with user interaction}.
\newblock \emph{Proceedings of the Web Conference 2021}.

\bibitem[{Gupta et~al.(2020)Gupta, Yang, Sivakumar, Silva, Hall, and Barioni}]{InstagramTemporal}
Aabhaas Gupta, Wenxia Yang, Divya Sivakumar, Yasin~N. Silva, Deborah~L. Hall, and Maria Camila~Nardini Barioni. 2020.
\newblock \href {https://api.semanticscholar.org/CorpusID:212412836} {Temporal properties of cyberbullying on instagram}.
\newblock \emph{Companion Proceedings of the Web Conference 2020}.

\bibitem[{Hamlett et~al.(2022)Hamlett, Powell, Silva, and Hall}]{MulitiLabelsIns}
Mara Hamlett, Grace Powell, Yasin~N. Silva, and Deborah~L. Hall. 2022.
\newblock \href {https://api.semanticscholar.org/CorpusID:248980841} {A labeled dataset for investigating cyberbullying content patterns in instagram}.
\newblock In \emph{International Conference on Web and Social Media}.

\bibitem[{Hartvigsen et~al.(2022)Hartvigsen, Gabriel, Palangi, Sap, Ray, and Kamar}]{TOXIGEN}
Thomas Hartvigsen, Saadia Gabriel, Hamid Palangi, Maarten Sap, Dipankar Ray, and Ece Kamar. 2022.
\newblock \href {https://api.semanticscholar.org/CorpusID:247519233} {Toxigen: A large-scale machine-generated dataset for adversarial and implicit hate speech detection}.
\newblock In \emph{Annual Meeting of the Association for Computational Linguistics}.

\bibitem[{Hochreiter and Schmidhuber(1997)}]{LSTM}
Sepp Hochreiter and J{\"u}rgen Schmidhuber. 1997.
\newblock \href {https://api.semanticscholar.org/CorpusID:1915014} {Long short-term memory}.
\newblock \emph{Neural Computation}, 9:1735--1780.

\bibitem[{Hosseinmardi et~al.(2015)Hosseinmardi, Mattson, Rafiq, Han, Lv, and Mishra}]{2015Instagram}
Homa Hosseinmardi, Sabrina~Arredondo Mattson, Rahat~Ibn Rafiq, Richard~O. Han, Qin Lv, and Shivakant Mishra. 2015.
\newblock \href {https://api.semanticscholar.org/CorpusID:1162787} {Analyzing labeled cyberbullying incidents on the instagram social network}.
\newblock In \emph{Social Informatics}.

\bibitem[{Huang et~al.(2023)Huang, Kwak, and An}]{GPTAnno}
Fan Huang, Haewoon Kwak, and Jisun An. 2023.
\newblock \href {https://api.semanticscholar.org/CorpusID:256868854} {Is chatgpt better than human annotators? potential and limitations of chatgpt in explaining implicit hate speech}.
\newblock \emph{Companion Proceedings of the ACM Web Conference 2023}.

\bibitem[{Jiang et~al.(2021)Jiang, Yang, Liu, and Zubiaga}]{SwSR}
Aiqi Jiang, Xiaohan Yang, Yang Liu, and Arkaitz Zubiaga. 2021.
\newblock \href {https://api.semanticscholar.org/CorpusID:236950681} {Swsr: A chinese dataset and lexicon for online sexism detection}.
\newblock \emph{Online Soc. Networks Media}, 27:100182.

\bibitem[{Jiawen et~al.(2022)Jiawen, Zhou, Sun, Zheng, Mi, and Huang}]{COLD}
Deng Jiawen, Jingyan Zhou, Hao Sun, Chujie Zheng, Fei Mi, and Minlie Huang. 2022.
\newblock \href {https://api.semanticscholar.org/CorpusID:246016271} {Cold: A benchmark for chinese offensive language detection}.
\newblock In \emph{Conference on Empirical Methods in Natural Language Processing}.

\bibitem[{Kim(2014)}]{CNN}
Yoon Kim. 2014.
\newblock \href {https://api.semanticscholar.org/CorpusID:9672033} {Convolutional neural networks for sentence classification}.
\newblock In \emph{Conference on Empirical Methods in Natural Language Processing}.

\bibitem[{Liu et~al.(2019)Liu, Ott, Goyal, Du, Joshi, Chen, Levy, Lewis, Zettlemoyer, and Stoyanov}]{RoBERTa}
Yinhan Liu, Myle Ott, Naman Goyal, Jingfei Du, Mandar Joshi, Danqi Chen, Omer Levy, Mike Lewis, Luke Zettlemoyer, and Veselin Stoyanov. 2019.
\newblock \href {https://api.semanticscholar.org/CorpusID:198953378} {Roberta: A robustly optimized bert pretraining approach}.
\newblock \emph{ArXiv}, abs/1907.11692.

\bibitem[{Lu et~al.(2023)Lu, Xu, Zhang, Min, Yang, and Lin}]{TOXICN}
Junyu Lu, Bo~Xu, Xiaokun Zhang, Chang~Hyo Min, Liang Yang, and Hongfei Lin. 2023.
\newblock \href {https://api.semanticscholar.org/CorpusID:258557119} {Facilitating fine-grained detection of chinese toxic language: Hierarchical taxonomy, resources, and benchmarks}.
\newblock \emph{ArXiv}, abs/2305.04446.

\bibitem[{Maity et~al.(2022)Maity, Jha, Saha, and Bhattacharyya}]{Maity2022AMF}
Krishanu Maity, Prince Jha, Sriparna Saha, and Pushpak Bhattacharyya. 2022.
\newblock \href {https://api.semanticscholar.org/CorpusID:248962585} {A multitask framework for sentiment, emotion and sarcasm aware cyberbullying detection from multi-modal code-mixed memes}.
\newblock \emph{Proceedings of the 45th International ACM SIGIR Conference on Research and Development in Information Retrieval}.

\bibitem[{Mathew et~al.(2020)Mathew, Saha, Yimam, Biemann, Goyal, and Mukherjee}]{DataImbalance}
Binny Mathew, Punyajoy Saha, Seid~Muhie Yimam, Chris Biemann, Pawan Goyal, and Animesh Mukherjee. 2020.
\newblock \href {https://api.semanticscholar.org/CorpusID:229332119} {Hatexplain: A benchmark dataset for explainable hate speech detection}.
\newblock In \emph{AAAI Conference on Artificial Intelligence}.

\bibitem[{Murshed et~al.(2022)Murshed, Abawajy, Mallappa, Saif, and Al-ariki}]{Murshed2022DEARNNAH}
Belal Abdullah~Hezam Murshed, Jemal~H. Abawajy, Suresha Mallappa, Mufeed Ahmed~Naji Saif, and Hasib Daowd~Esmail Al-ariki. 2022.
\newblock \href {https://api.semanticscholar.org/CorpusID:247102635} {Dea-rnn: A hybrid deep learning approach for cyberbullying detection in twitter social media platform}.
\newblock \emph{IEEE Access}, 10:25857--25871.

\bibitem[{Rafiq et~al.(2015)Rafiq, Hosseinmardi, Han, Lv, Mishra, and Mattson}]{2015Vine}
Rahat~Ibn Rafiq, Homa Hosseinmardi, Richard~O. Han, Qin Lv, Shivakant Mishra, and Sabrina~Arredondo Mattson. 2015.
\newblock \href {https://api.semanticscholar.org/CorpusID:12841728} {Careful what you share in six seconds: Detecting cyberbullying instances in vine}.
\newblock \emph{2015 IEEE/ACM International Conference on Advances in Social Networks Analysis and Mining (ASONAM)}, pages 617--622.

\bibitem[{Rosa et~al.(2018)Rosa, Carvalho, Calado, Martins, Ribeiro, and Coheur}]{10}
Hugo Rosa, Joao~P. Carvalho, Pável Calado, Bruno Martins, Ricardo Ribeiro, and Luisa Coheur. 2018.
\newblock \href {https://doi.org/10.1109/FUZZ-IEEE.2018.8491557} {Using fuzzy fingerprints for cyberbullying detection in social networks}.
\newblock In \emph{2018 IEEE International Conference on Fuzzy Systems (FUZZ-IEEE)}, pages 1--7.

\bibitem[{Royen et~al.(2017)Royen, Poels, Vandebosch, and Adam}]{9}
Kathleen~Van Royen, Karolien Poels, Heidi Vandebosch, and Philippe C~G Adam. 2017.
\newblock \href {https://api.semanticscholar.org/CorpusID:25230057} {"thinking before posting?" reducing cyber harassment on social networking sites through a reflective message}.
\newblock \emph{Comput. Hum. Behav.}, 66:345--352.

\bibitem[{Salawu et~al.(2020)Salawu, He, and Lumsden}]{77}
Semiu Salawu, Yulan He, and Joan~A. Lumsden. 2020.
\newblock \href {https://api.semanticscholar.org/CorpusID:59134249} {Approaches to automated detection of cyberbullying: A survey}.
\newblock \emph{IEEE Transactions on Affective Computing}, 11:3--24.

\bibitem[{Smith et~al.(2008)Smith, Mahdavi, de~Carvalho, Fisher, Russell, and Tippett}]{CB-repeat}
Peter~K. Smith, Jessica Mahdavi, MD. Manuel~H. de~Carvalho, Sonja Fisher, Shanette Russell, and Neil Tippett. 2008.
\newblock \href {https://api.semanticscholar.org/CorpusID:28942016} {Cyberbullying: its nature and impact in secondary school pupils.}
\newblock \emph{Journal of child psychology and psychiatry, and allied disciplines}, 49 4:376--85.

\bibitem[{Soni and Singh(2018)}]{Time2}
Devin Soni and Vivek~K. Singh. 2018.
\newblock \href {https://api.semanticscholar.org/CorpusID:49408114} {Time reveals all wounds: Modeling temporal characteristics of cyberbullying}.
\newblock In \emph{International Conference on Web and Social Media}.

\bibitem[{Sun et~al.(2019)Sun, Qiu, Xu, and Huang}]{transaction}
Chi Sun, Xipeng Qiu, Yige Xu, and Xuanjing Huang. 2019.
\newblock \href {https://api.semanticscholar.org/CorpusID:153312532} {How to fine-tune bert for text classification?}
\newblock In \emph{China National Conference on Chinese Computational Linguistics}.

\bibitem[{Wang et~al.(2020)Wang, Fu, and Lu}]{34}
Jason Wang, Kaiqun Fu, and Chang-Tien Lu. 2020.
\newblock \href {https://api.semanticscholar.org/CorpusID:232373954} {Sosnet: A graph convolutional network approach to fine-grained cyberbullying detection}.
\newblock \emph{2020 IEEE International Conference on Big Data (Big Data)}, pages 1699--1708.

\bibitem[{Wulczyn et~al.(2016)Wulczyn, Thain, and Dixon}]{32}
Ellery Wulczyn, Nithum Thain, and Lucas Dixon. 2016.
\newblock \href {https://api.semanticscholar.org/CorpusID:6060248} {Ex machina: Personal attacks seen at scale}.
\newblock \emph{Proceedings of the 26th International Conference on World Wide Web}.

\bibitem[{Yi and Zubiaga(2022)}]{Yi2022SessionbasedCD}
Peiling Yi and Arkaitz Zubiaga. 2022.
\newblock \href {https://api.semanticscholar.org/CorpusID:250921165} {Session-based cyberbullying detection in social media: A survey}.
\newblock \emph{Online Soc. Networks Media}, 36:100250.

\bibitem[{Yi and Zubiaga(2023)}]{Yi2023LearningLH}
Peiling Yi and Arkaitz Zubiaga. 2023.
\newblock \href {https://api.semanticscholar.org/CorpusID:258333835} {Learning like human annotators: Cyberbullying detection in lengthy social media sessions}.
\newblock \emph{Proceedings of the ACM Web Conference 2023}.

\bibitem[{Zhang et~al.(2016)Zhang, Tong, Vishwamitra, Whittaker, Mazer, Kowalski, Hu, Luo, Macbeth, and Dillon}]{imblance2}
Xiang Zhang, Jonathan Tong, Nishant Vishwamitra, Elizabeth Whittaker, Joseph~P. Mazer, Robin~M. Kowalski, Hongxin Hu, Feng Luo, Jamie~C. Macbeth, and Edward~C. Dillon. 2016.
\newblock \href {https://api.semanticscholar.org/CorpusID:206823112} {Cyberbullying detection with a pronunciation based convolutional neural network}.
\newblock \emph{2016 15th IEEE International Conference on Machine Learning and Applications (ICMLA)}, pages 740--745.

\end{thebibliography}

\clearpage
\appendix

\section{Supplement of Dataset Description}
\label{sec:A}

\subsection{Source Composition of Sessions}
\label{sec:source}

Our data were collected through three different strategies. Here, we provide an overview of the data distribution associated with each strategy. Although we do not have precise statistical data, based on our estimates, keyword-related data account for over 30\%, data related to major events account for over 20\%, and data related to everyday popular events account for over 40\%.

\subsection{Details of Dataset}
To enable researchers to understand and utilize our dataset, we provide a detailed description of its components. The dataset consists of four parts: post information, comment information, repost information, and user details.

\subsubsection{Posts}
\begin{figure}[H]
  \includegraphics[width=\columnwidth]{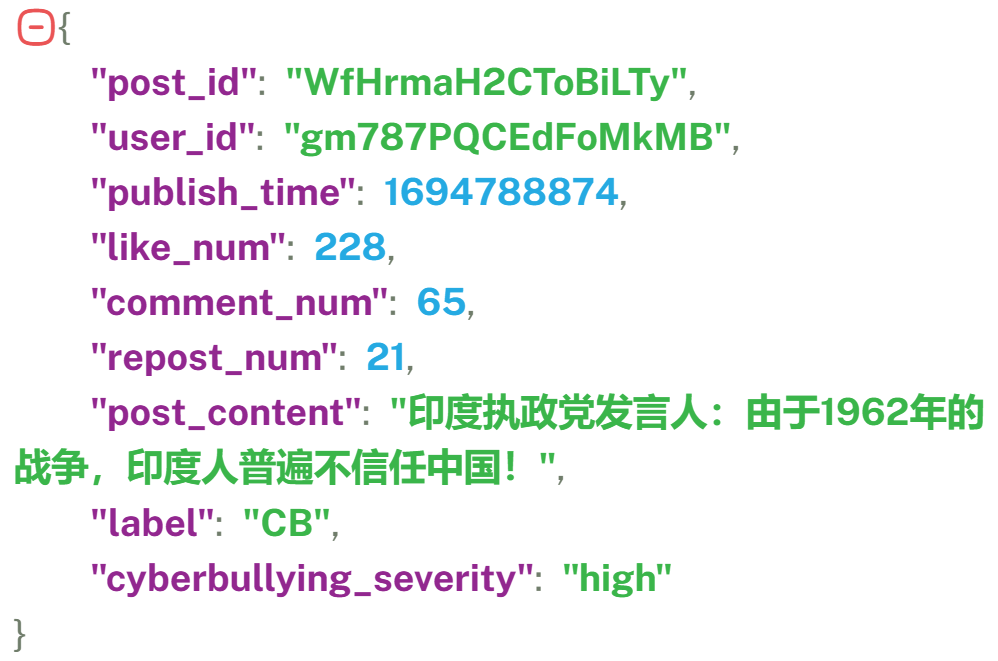}
  \caption{A sample of post information in the dataset.}
  \label{fig:post}
\end{figure}

The original post functions as both the starting point and the central content of the entire session. Consequently, we offer detailed information for each post, including the poster’s ID, posting time, number of likes, number of comments, number of reposts, and the content of the post. A sample is shown in Figure \ref{fig:post}.

To address the impact of time zone discrepancies, we standardize posting times using UNIX timestamps. It is important to note that the number of comments or reposts we have stored may be less than the numbers reported here, as some contents may have been deleted or blocked.

\subsubsection{Comments}
\begin{figure}[H]
  \includegraphics[width=0.87\columnwidth]{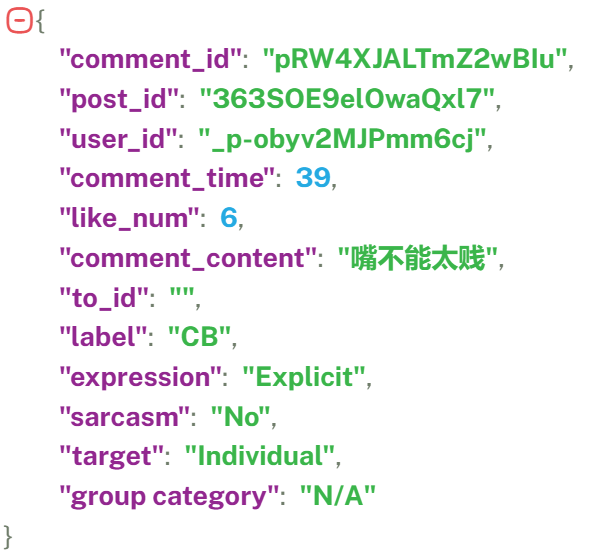}
  \caption{An example of comment information in the dataset.}
  \label{fig:comment}
\end{figure}

In our dataset, we provide information related to comments, including comment ID, post ID, user ID, comment time, number of likes, comment content, ID of the replied comment, and five annotated labels. An example is presented in Figure \ref{fig:comment}.

To further investigate the temporal properties of cyberbullying, the comment time is recorded as the difference from the original post's timestamp, measured in minutes. If a comment is not a reply to another comment, the to\_id field remains empty.

\subsubsection{Reposts}
\begin{figure}[H]
  \includegraphics[width=0.8\columnwidth]{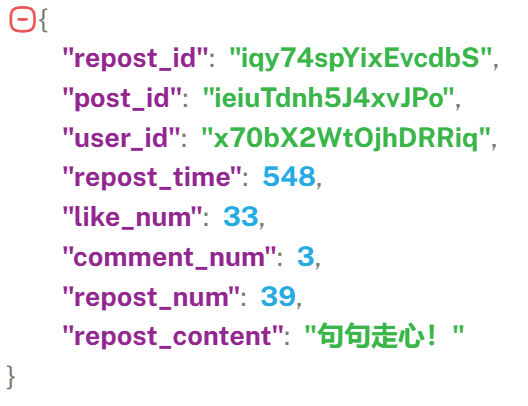}
  \caption{A sample of repost in the dataset.}
  \label{fig:repost}
\end{figure}

Similar to Twitter, Weibo allows users to repost others' posts with additional comments. Posts created through reposting are commonly referred to as "quote posts". We also provide information on quote posts within the sessions, as this form of content dissemination can potentially initiate or escalate cyberbullying incidents. In Figure \ref{fig:repost}, we present an illustration of repost.

The handling of repost time follows the same approach as comment timestamps, being converted into time differences. In addition, if a user reposts a Weibo without adding a comment, the platform automatically sets the repost content to "Repost Weibo." In such cases, we remove the repost content.

\subsubsection{User Details}
\begin{figure}[H]
  \includegraphics[width=0.75\columnwidth]{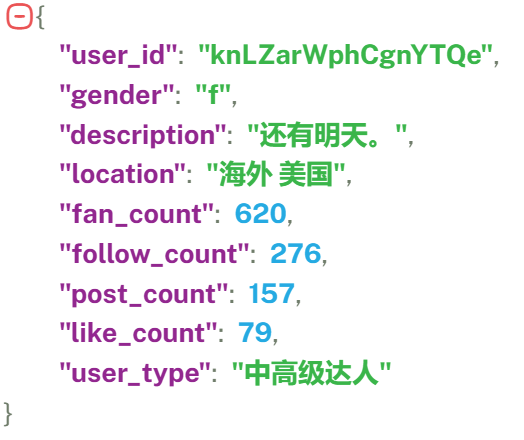}
  \caption{Overview of all stored information for a user in the dataset.}
  \label{fig:user}
\end{figure}

The dataset includes information on three types of users: posters, commenters, and reposters. The user information includes user ID, gender, personal description, location, number of followers, number of friends, number of posts, number of likes received, and user type. Here we provide an example in Figure \ref{fig:user}.

\section{Keywords of Data Collection}
\label{sec:keywords}

The keywords of each topic used in data collection are shown in Table \ref{table key}.

\begin{table}[H]

  \centering
  \renewcommand\arraystretch{1.5}
  \begin{tabular}{  m{0.08\textwidth}  m{0.35\textwidth}}
    \toprule

     \textbf{Topic} & \textbf{Keywords}   \\
    \midrule
        Sexism & \begin{CJK*}{UTF8}{gbsn} 性别平等, 性别歧视, 两性, 婚姻, 女性, 妇女, 彩礼  \end{CJK*} \\

        Region & \begin{CJK*}{UTF8}{gbsn}  地域黑, 河南人, 东北人, 南北方, 农村, 外省, 文化差异, 上海排外, 洋垃圾, 小日本, 韩国, 棒子 \end{CJK*} \\
        
        Racism & \begin{CJK*}{UTF8}{gbsn} 黑鬼, 黑人, 白人, 白皮, 印度人, 黄种人  \end{CJK*}  \\

        LGBTQ & \begin{CJK*}{UTF8}{gbsn} lgbt, 男同, 女同, 双性恋, 跨性别, 性少数  \end{CJK*} \\

    \bottomrule
  \end{tabular}
  \caption{\label{table key}
    Topic and keywords.
  }
\end{table}

\section{Details of Annotation}
\label{sec:C}

\subsection{Demonstration}
\begin{figure}[H]
  \centering
  \includegraphics[width=\columnwidth]{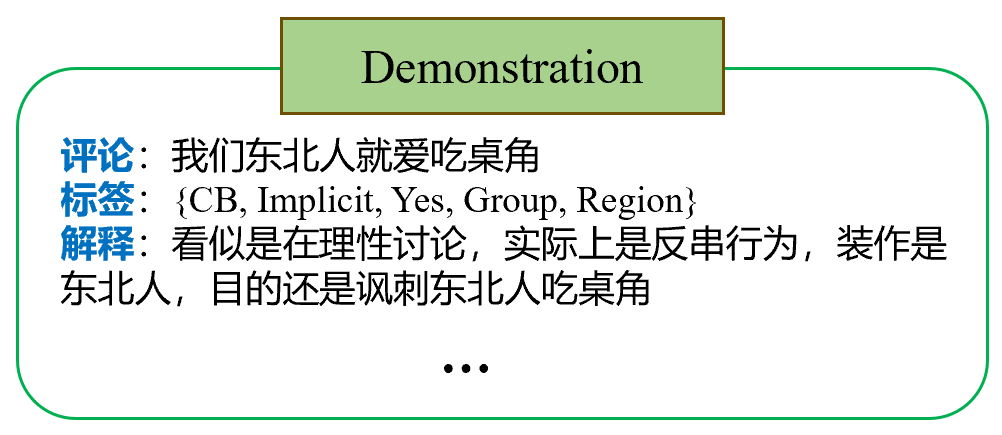}
  \caption{A demonstration from the annotation process, which includes a comment, five lables and an explanation.}
  \label{fig:demons}
\end{figure}

 When utilizing the LLM for annotation, we employ demonstration-based prompting to guide the process. A sample of demonstration is shown in Figure \ref{fig:demons}.

To ensure accuracy, each example is thoroughly discussed, and the final label is jointly determined by all annotators. What's more, the core of the demonstration lies in providing detailed explanations to help annotators clearly understand the meaning of complex comments. The primary goal is to assist the LLM in accurately labeling comments that are highly ambiguous or polysemous, thereby avoiding misunderstandings or biases. This process enhances the overall accuracy and consistency of the annotations.




\subsection{Annotation Guideline}

\begin{figure}[H]
  \centering
  \includegraphics[width=\columnwidth]{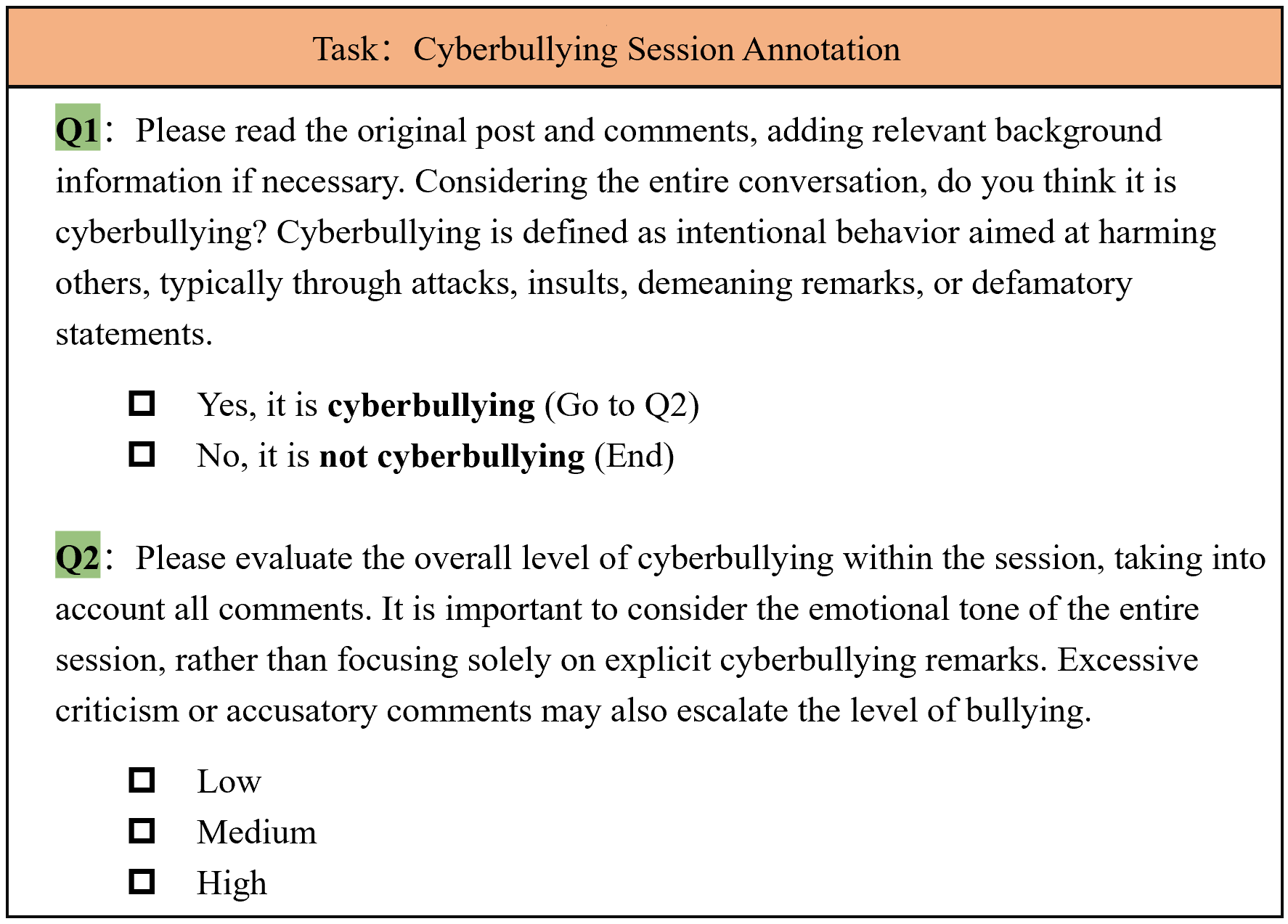}
  \caption{ Summary of our session annotation guidelines.}
  \label{fig:session_guide}
\end{figure}

The annotation guideline of comments is presented in Figure \ref{fig:guide}. Therefore, we provide the annotation guide for sessions, as shown in Figure \ref{fig:session_guide}. It is important to note that a session will not be labeled as cyberbullying simply because it contains a few cyberbullying comments. A session is only classified as cyberbullying when the amount of cyberbullying content reaches a certain threshold.

\subsection{Prompt of Role Definition}
\label{sec:role}

In this section, we provide the detailed prompts designed within the role definition. The special prompt is as follows:

\begin{quote}
    \textit{You are an expert in Chinese language analysis and a seasoned internet user deeply familiar with online culture and communication styles. You excel at identifying bullying behavior in online interactions, including explicit and implicit expressions, as well as sarcasm or humor that may disguise aggressive language.}
\end{quote}

\end{document}